\documentclass[conference,a4paper]{IEEEtran}
\IEEEoverridecommandlockouts

\usepackage{amsmath,amssymb,amsfonts,amsthm}
\usepackage[ruled,vlined]{algorithm2e}
\usepackage{enumerate}
\usepackage{graphicx}
\usepackage{textcomp}
\usepackage{psfrag}
\usepackage{makecell}
\usepackage{url}
\usepackage[all]{xy}
\usepackage{multirow}
\usepackage{hyperref}
\def\BibTeX{{\rm B\kern-.05em{\sc i\kern-.025em b}\kern-.08em
    T\kern-.1667em\lower.7ex\hbox{E}\kern-.125emX}}

\newcommand{\R}{\mathbb{R}}

\newcommand{\A}{\mathbb{A}}

\newcommand{\bb}{\begin{equation}}
\newcommand{\ee}{\end{equation}}
\newcommand{\bbb}{\begin{eqnarray}}
\newcommand{\eee}{\end{eqnarray}}
\newcommand{\benu}{\begin{enumerate}}
\newcommand{\eenu}{\end{enumerate}}

\newcommand{\bpm}{\begin{bmatrix}}
\newcommand{\epm}{\end{bmatrix}}

\newcommand{\ii}{\boldsymbol{i}}
\newcommand{\jj}{\boldsymbol{j}}
\newcommand{\kk}{\boldsymbol{k}}

\newcommand{\quat}[1]{{#1}_0 + {#1}_1 \ii + {#1}_2 \jj + {#1}_3 \kk}

\newcommand{\hyperN}[1]{{#1}_0 + {#1}_1 \boldsymbol{i}_{1} +  \dots + {#1}_n \boldsymbol{i}_{n}}

\def\boxmax{\kern 0em\hbox{\rm \kern .25em\lower.1ex\hbox{\rlap{$\vee$}}\kern -.07em\lower.2ex\hbox{$\square$}\kern.25em}}
\def\boxmin{\kern 0em\hbox{\rm \kern .25em\lower.1ex\hbox{\rlap{$\wedge$}}\kern -.07em\lower.2ex\hbox{$\square$}\kern.25em}}
\def\boxdiamond{\kern 0em\hbox{\rm \kern .25em\hbox{\rlap{$\diamond$}}\kern -.15em\lower.2ex\hbox{$\square$}}}

\theoremstyle{definition}

\newtheorem{remark}{Remark}

\begin{document}
\newpage
\thispagestyle{empty}
\begin{minipage}{0.8\textwidth}
\noindent {\Huge  IEEE Copyright Notice} \vspace{2cm}

\noindent \copyright 2022 IEEE.  Personal use of this material is permitted.  Permission from IEEE must be obtained for all other uses, in any current or future media, including reprinting/republishing this material for advertising or promotional purposes, creating new collective works, for resale or redistribution to servers or lists, or reuse of any copyrighted component of this work in other works.
\vspace{1cm}

\noindent Accepted to be Published in: Proceedings of the 2022 International Joint Conference on Neural Networks (IJCNN 2022), 18-23 July, 2022, Padua, Italy.
\end{minipage}

\title{Acute Lymphoblastic Leukemia Detection Using Hypercomplex-Valued Convolutional Neural Networks 
\thanks{This work was supported in part by CNPq under grant no. 315820/2021-7, FAPESP under grant no. 2019/02278-2, and Coordena\c{c}\~ao  de Aperfei\c{c}oamento de Pessoal de N\'ivel Superior - Brasil (CAPES) - Finance Code 001.}
}

\author{
\IEEEauthorblockN{1\textsuperscript{st} Guilherme Vieira}
\IEEEauthorblockA{\textit{Department of Applied Mathematics} \\
\textit{University of Campinas}\\
Campinas, Brazil \\
email: vieira.g@ime.unicamp.br}
\and
\IEEEauthorblockN{2\textsuperscript{nd} Marcos Eduardo Valle}
\IEEEauthorblockA{\textit{Department of Applied Mathematics} \\
\textit{University of Campinas}\\
Campinas, Brazil \\
email: valle@ime.unicamp.br}
}




\maketitle

\begin{abstract}
This paper features convolutional neural networks defined on hypercomplex algebras applied to classify lymphocytes in blood smear digital microscopic images. Such classification is helpful for the diagnosis of acute lymphoblast leukemia (ALL), a type of blood cancer. We perform the classification task using eight hypercomplex-valued convolutional neural networks (HvCNNs) along with real-valued convolutional networks. Our results show that HvCNNs perform better than the real-valued model, showcasing higher accuracy with a much smaller number of parameters. Moreover, we found that HvCNNs based on Clifford algebras processing HSV-encoded images attained the highest observed accuracies. Precisely, our HvCNN yielded an average accuracy rate of 96.6\% using the ALL-IDB2 dataset with a 50\% train-test split, a value extremely close to the state-of-the-art models but using a much simpler architecture with significantly fewer parameters.
\end{abstract}


\section{Introduction}

Over the past few decades, artificial neural networks (ANN) have been employed on various classification tasks, many of them previously performed by humans. One popular example is the computer-assisted diagnosis (CAD), in which the output of the ANN may assist doctors in making more accurate decisions \cite{eadie2012systematic}. Most applications in CAD consist of a machine learning model performing tasks that would be handmade by specialists, reducing the financial and human costs while also avoiding possible mistakes caused by fatigue \cite{pereira2019survey,qian1995computer}. Moreover, computers have been repeatedly appointed as outperforming unaided professionals in these tasks \cite{qian1995computer,leaper1972computer}. CAD applications have been effectively used, for example, for leukemia identification \cite{Salah2019MachineDirections}. 

Acute lymphoblastic leukemia (ALL) is a blood pathology that can be lethal in only a few weeks if left unchecked. ALL is a type of blood cancer identified by immature lymphocytes, known as lymphoblasts, in the blood and bone marrow. The peak incidence lies in children ages 2-5 years, and one of the primary forms of ALL detection is through microscopic blood sample inspection. Computer-aided leukemia diagnosis has been achieved using many different approaches in the past years, including deep learning models combined with transfer learning and unsharpening techniques \cite{Bibi2020IoMT-Based,Genovese2021HistopathologicalDetection,Genovese2021AcuteLearning,Zolfaghari2022ACells}. Also, it has been reported in the literature that a quaternion-valued convolutional neural network exhibited better generalization capability than its corresponding real-valued model to classify white cells as lymphoblasts \cite{Granero2021Quaternion-ValuedDiagnosis}. Furthermore, the quaternion-valued CNN has about 34\% of the total number of parameters of the corresponding real-valued network. This paper investigates further the application of hypercomplex-valued convolutional neural networks (HvCNN) ALL detection.

Like complex numbers, quaternions are hypercomplex numbers with a wide range of applications in science and engineering. For example, quaternions provide an efficient tool for describing 3D rotations used in computer graphics. Furthermore, quaternion-valued neural networks have been effectively applied for signal processing and control \cite{Talebi2020Quaternion-ValuedControl,Takahashi2021ComparisonControl}, image classification \cite{shang14,Valle2020Quaternion-valuedQuaternions}, and many other pattern recognition tasks \cite{Parcollet2020ANetworks,Bayro-Corrochano2020QuaternionApplications}. However, besides the quaternions, many other hypercomplex-valued algebras exist, including the coquaternions, the tessarines, Clifford algebras, and Cayley-Dickson algebras. The tessarines, introduced by Cockle a few years after the introduction of quaternions, is a commutative hypercomplex algebra that has been effectively used for signal processing and the development of neural networks \cite{Navarro-Moreno2020TessarineCondition,Navarro-Moreno2021Wide-senseConditions,Ortolani2017OnProcessing,Carniello2021UniversalApproximationTessarineNetworks,Senna2021TessarineQuaternionImageClassification}. The Clifford algebras comprise a broad family of associative hypercomplex algebras with interesting geometrical properties \cite{hestenes12,hitzer13,vaz16}. A family of hypercomplex algebras, obtained from the recursive process of Cayley-Dickson, can be used to develop effective HvNNs \cite{Brown1967OnAlgebras.,Vieira2020ExtremeAuto-Encoding}. Furthermore, four-dimensional hypercomplex algebras have been used for designing neural networks for controlling a robot manipulator \cite{Takahashi2021ComparisonControl}.

This paper considers the hypercomplex algebra framework proposed by Kantor and Solodovnikov \cite{Kantor1989HypercomplexAlgebras}. This approach furnishes a broad class of hypercomplex algebras which comprises the tessarines, the Clifford algebras, and the Cayley-Dickson algebras. On the downside, the general framework by Kantor and Solodovnikov also contains many hypercomplex algebras with no attractive features. Thus, we shall focus only on eight four-dimensional associative hypercomplex algebras, four of which are commutative. We consider the tessarines, the bi-complex numbers, and the Klein 4-group algebra among the commutative algebras \cite{Takahashi2021ComparisonControl,Kobayashi2020HopfieldFour-group}. The four non-commutative hypercomplex algebras include quaternions and coquaternions. Also, they are isomorphic to Clifford as well as Cayley-Dickson algebras \cite{Vieira2020ExtremeAuto-Encoding,Vieira2022AMachines}.

The paper is organized as follows: Next section presents the basic concepts of hypercomplex algebras and explores notable four-dimensional algebras. Section \ref{sec:hvcnn} addresses hypercomplex-valued CNN models and shows how to emulate them using a real-valued convolutional network. Section \ref{sec:Applications} describes the computational experiments conducted using the ALL-IDB dataset. The paper finishes with some concluding remarks in Section \ref{sec:concluding-remarks}.

\section{Hypercomplex Numbers and Some Notable Associative Four-Dimensional Algebras} \label{sec:hc-matrix-algebra}

Let us present a few basic definitions regarding hypercomplex numbers. Hypercomplex algebras can be defined in any field, but we focus on algebras over the real numbers in this work. As pointed out in the introduction, we consider the general framework proposed by Kantor and Solodovnikov, which includes the most used hypercomplex algebras \cite{Kantor1989HypercomplexAlgebras}. We finish this section by presenting eight notable associative four-dimensional hypercomplex algebras.


\subsection{A Brief Review on Hypercomplex Number Systems} \label{sec:review}

A hypercomplex algebra $\A$ of dimension $n+1$ over $\R$ is a set
\begin{equation}
\A = \{ x = x_0 + x_1 \boldsymbol{i}_1 + x_2 \boldsymbol{i}_2 + \dots + x_n \boldsymbol{i}_n: x_\mu \in \R, \forall \mu \},
\end{equation}
equipped with two operations, namely addition and multiplication \cite{Kantor1989HypercomplexAlgebras}. The symbols $\ii_\mu$, with $\ii_\mu \notin \R$ for all $\mu = 1,\ldots,n$, denote the so-called hyperimaginary units of $\A$. 
The addition of two hypercomplex numbers $x = \hyperN{x}$ and $y = \hyperN{y}$ is simply defined as
\begin{equation}
    \label{eq:addition}
    x + y = (x_0+y_0)+(x_1+y_1)\boldsymbol{i}_1+\ldots+(x_n+y_n)\boldsymbol{i}_n.
\end{equation} The multiplication operation is carried out distributively and replacing the product of two hypercomplex units $\ii_\mu$ and $\ii_\nu$ by an hypercomplex number $p_{\mu\nu} := \ii_\mu \ii_\nu$, where
\begin{equation} \label{eq:multiplication_rule}
p_{\mu\nu} = \hyperN{(p_{\mu\nu})} \in \mathbb{A}, 
\end{equation}
for all $\mu,\nu=1,\ldots,n$. Precisely, the multiplication of two hypercomplex numbers is given by the equation
\begin{equation}
\begin{aligned} \label{eq:multiplication}
xy &= \left(x_0 y_0 + \sum_{\mu,\nu=1}^n x_\mu y_\nu (p_{\mu\nu})_0 \right) \\  & 
+ \left(x_0y_1 + x_1 y_0+  
 \sum_{\mu,\nu=1}^n x_\mu y_\nu (p_{\mu\nu})_1 \right) \ii_1 + \ldots \\ &+ \left( x_0 y_n + x_ny_0 +  \sum_{\mu,\nu=1}^n x_\mu y_\nu (p_{\mu\nu})_n \right)\ii_n.
\end{aligned}
\end{equation}

Note that hyperimaginary units products characterize a hypercomplex algebra $\A$. In other words, the hypercomplex numbers $p_{\mu\nu}$ given by \eqref{eq:multiplication_rule} determine $\A$. It is common practice to arrange the hypercomplex numbers $p_{\mu\nu}$ in a table, called the multiplication table. Examples of multiplication tables are given in Tables \ref{tab:four-dimension-alg} and \ref{tab:non_cd_algebras}. Algebraic properties of an hypercomplex algebra $\mathbb{A}$ can be inferred from its multiplication tables. For example, symmetric multiplication tables represent commutative hypercomplex algebras \cite{Kantor1989HypercomplexAlgebras}. Note that the four multiplication tables in Table \ref{tab:non_cd_algebras} are symmetric. Thus, they represent commutative hypercomplex algebras. We will discuss in detail the multiplication tables provided in Tables \ref{tab:four-dimension-alg} and \ref{tab:non_cd_algebras} in the following subsection.

A scalar $\alpha \in \A$ can be identified with the hypercomplex number $\alpha = \alpha + 0 \ii_1 + \dots + 0 \ii_n$, and vice-versa. Furthermore, from \eqref{eq:multiplication}, the product by scalar in any algebra $\A$ satisfies
\begin{equation} \label{eq:scalar_product}
\alpha x = \hyperN{\alpha x}.
\end{equation}
This remark shows that a hypercomplex algebra $\A$ can be identified with an $(n+1)$-dimensional vector space with the vector addition and the product by a scalar given by \eqref{eq:addition} and \eqref{eq:scalar_product}, respectively. 
Moreover, the basis elements of such $(n+1)$-dimensional vector space are $1,\ii_1,\ldots,\ii_n$. Besides the addition and the product by a scalar, a hypercomplex algebra is equipped with the multiplication of vectors given by \eqref{eq:multiplication}.

We would like to point out that different multiplication tables do not necessarily yield different hypercomplex algebras. Precisely,
we know from linear algebra that a vector space can be represented using different bases. Similarly, a hypercomplex algebra can be obtained from different bases or, equivalently, using different hypercomplex units. Since the multiplication table determines the outcome of the product of any two hypercomplex units, a change of basis results in a different multiplication table. Because of this remark, we say  two hypercomplex algebras $\A$ and $\A'$ are isomorphic if they differ by a change of basis. In other words, $\A$ and $\A'$ are isomorphic hypercomplex algebras if the multiplication table of $\A'$ can be obtained from the multiplication table of $\A$ through a change of basis.



\subsection{Four-dimensional Hypercomplex Algebras} \label{sec:4dim}

Let us now focus on four-dimensional hypercomplex algebras, i.e., hypercomplex algebras of the form 
\begin{equation} \label{eq:4D}
\A = \{x = \quat{x}: x_0,\ldots,x_3 \in \R\},
\end{equation}
where $\ii \equiv \ii_1$, $\jj \equiv \ii_2$, and $\kk \equiv \ii_3$ are the three hypercomplex units. Four-dimensional hypercomplex algebras are particularly useful for image processing because a color can be represented using a single hypercomplex number. In other words, four-dimensional hypercomplex algebras allows representing a color by a single entity. 

Note that the hypercomplex numbers $p_{\mu\nu} = \quat{(p_{\mu\nu})}$ in the multiplication table results in a large number of algebras. Some of these algebras are isomorphic, and many of them have no attractive features. Therefore, we present a construction rule that yields eight associative four-dimensional hypercomplex algebras in the following.

Like in the construction of Clifford algebras \cite{Renaud2020CLIFFORDPHYSICS}, we assume the product of hypercomplex units is associative. Furthermore, we let the identity 
\begin{equation} \label{eq:k=ij}
\kk = \ii \jj,
\end{equation} hold true. Finally, we assume the four-dimensional hypercomplex algebra is either commutative or anti-commutative. 




\subsubsection{Anticommutative Algebras} 
We obtain an anticommutative four-dimensional hypercomplex algebra by imposing $\ii \jj = -\jj \ii$. From the associativity and \eqref{eq:k=ij}, we obtain
\begin{align}
\kk^2 &= (\ii\jj)(\ii\jj) = \ii (\jj \ii) \jj = \ii (-\ii \jj) \jj = - \ii^2 \jj^2, \label{eq:A-k2}\\
\ii \kk &= \ii (\ii\jj) = \ii^2 \jj, \\
\jj \kk &= \jj (\ii\jj) = \jj (-\jj\ii) = - \jj^2 \ii. \label{eq:A-jk}
\end{align}
To simplify the exposition, let $\ii^2 = \gamma_1$ and $\jj^2 = \gamma_2$. From \eqref{eq:k=ij}-\eqref{eq:A-jk}, we obtain an associative and anticommutative four-dimensional algebra denoted by $A[\gamma_1,\gamma_2]$, whose multiplication table is then
\begin{equation}
\begin{tabular}{c|rrr} 
    $A[\gamma_1,\gamma_2]$ & $\ii$ & $\jj$ & $\kk$ \\ \hline
    $\ii$ & $\gamma_1$ & $\kk$ & $\gamma_1 \jj$ \\  
    $\jj$ & $-\kk$ & $\gamma_2$ & $-\gamma_2 \ii$ \\  
    $\kk$ & $-\gamma_1 \jj$ & $\gamma_2 \ii$ & $-\gamma_1\gamma_2$  
\end{tabular} 
\end{equation}
By considering $\gamma_1,\gamma_2 \in \{-1,+1\}$, we obtain the four-dimensional hypercomplex algebras $A[-1,-1]$, $A[-1,+1]$, $A[+1,-1]$, and $A[+1,+1]$ whose multiplication tables are depicted in Table \ref{tab:four-dimension-alg}.

\begin{table*}[t]
\centering
    \caption{Multiplication tables of the anticommutative algebras.}
    \label{tab:four-dimension-alg}
\begin{tabular}{cccc}
    Quaternions & $C\ell_{2,0}$ & Coquaternions & $C\ell_{1,1}$ \\ 
\begin{tabular}{c|rrr} 
             $A[-1,-1]$ & $\ii$ & $\jj$ & $\kk$ \\ \hline
            $\ii$ & $-1$ & $\kk$ & $-\jj$ \\  
            $\jj$ & $-\kk$ & $-1$ & $\ii$ \\  
            $\kk$ & $\jj$ & $-\ii$ & $-1$\\  
        \end{tabular}  &  
\begin{tabular}{c|rrr} 
         $A\left[+1,+1\right]$ &  $\ii$ & $\jj$ & $\kk$ \\ \hline
         $\ii$ & $1$ & $\kk$ & $\jj$ \\  
         $\jj$ & $-\kk$ & $1$ & $-\ii$ \\ 
         $\kk$ & $-\jj$ & $\ii$ & $-1$\\  
    \end{tabular}   & 
\begin{tabular}{c|rrr}
     $A\left[-1,+1 \right]$ & $\ii$ & $\jj$ & $\kk$ \\ \hline
     $\ii$ & $-1$ & $\kk$ & $-\jj$ \\  
     $\jj$ & $-\kk$ & $1$ & $-\ii$ \\  
     $\kk$ & $\jj$ & $\ii$ & $1$\\  
    \end{tabular}
    &  
\begin{tabular}{c|rrr}
         $A\left[ +1,-1 \right]$ & $\ii$ & $\jj$ & $\kk$ \\ \hline
         $\ii$ & $1$ & $\kk$ & $\jj$ \\  
         $\jj$ & $-\kk$ & $-1$ & $\ii$ \\  
         $\kk$ & $-\jj$ & $-\ii$ & $1$\\  
    \end{tabular}  
\end{tabular}
\end{table*}  

\begin{table*}[t]
\centering
    \caption{Multiplication tables of commutative algebras.}
    \label{tab:non_cd_algebras}
\begin{tabular}{cccc}
Bicomplex numbers & & Tessarines & Klein 4-group \\
    \begin{tabular}{c|rrr} 
             $B[-1,-1]$ & $\ii$ & $\jj$ & $\kk$ \\ \hline
            $\ii$ & $-1$ & $\kk$ & $-\jj$ \\  
            $\jj$ & $\kk$ & $-1$ & $-\ii$ \\  
            $\kk$ & $-\jj$ & $-\ii$ & $1$\\  
        \end{tabular} & 
    \begin{tabular}{c|rrr} 
             $B[+1,-1]$ & $\ii$ & $\jj$ & $\kk$ \\ \hline
            $\ii$ & $1$ & $\kk$ & $\jj$ \\  
            $\jj$ & $\kk$ & $-1$ & $-\ii$ \\  
            $\kk$ & $\jj$ & $-\ii$ & $-1$\\  
        \end{tabular} &
    \begin{tabular}{c|rrr} 
             $B[-1,+1]$ & $\ii$ & $\jj$ & $\kk$ \\ \hline
            $\ii$ & $-1$ & $\kk$ & $-\jj$ \\  
            $\jj$ & $\kk$ & $1$ & $\ii$ \\  
            $\kk$ & $-\jj$ & $\ii$ & $-1$\\  
        \end{tabular} 
        &
    \begin{tabular}{c|rrr} 
             $B[+1,+1]$ & $\ii$ & $\jj$ & $\kk$ \\ \hline
            $\ii$ & $1$ & $\kk$ & $\jj$ \\  
            $\jj$ & $\kk$ & $1$ & $\ii$ \\  
            $\kk$ & $\jj$ & $\ii$ & $1$\\  
        \end{tabular} 
\end{tabular} 
\end{table*} 

\begin{remark}
The hypercomplex algebra $A[-1,-1]$ coincides with the quaternions because they have the same multiplication table. The algebra $A[-1,+1]$ corresponds to the co-quaternions, also known as split-quaternions \cite{Takahashi2021ComparisonControl}. Similarly, $A[+1,+1]$ and $A[+1,-1]$ can be identified with the Clifford algebras $C\ell_{2,0}$ and $C\ell_{1,1}$, respectively. Furthermore, the algebras $A[-1,+1]$, $A[+1,-1]$, and $A[+1,+1]$ are all isomorphic. Finally, we would like to remark that the algebras $A[-1,-1]$, $A[-1,+1]$, $A[+1,-1]$, and $A[+1,+1]$ can be derived using the generalized Cayley-Dickson process \cite{Brown1967OnAlgebras.,Vieira2020ExtremeAuto-Encoding}. 
\end{remark}

\subsubsection{Commutative Algebras}
In a similar fashion, we impose the condition $\ii \jj = \jj \ii$ to obtain the commutative four-dimensional hypercomplex algebras. Again, using associativity and \eqref{eq:k=ij}, we are able to write the identities
\begin{align}
\kk^2 &= (\ii\jj)(\ii\jj) = \ii (\jj \ii) \jj = \ii (\ii \jj) \jj = \ii^2 \jj^2, \label{eq:B-k2}\\
\ii \kk &= \ii (\ii\jj) = \ii^2 \jj, \\
\jj \kk &= \jj (\ii\jj) = \jj (\jj\ii) = \jj^2 \ii. \label{eq:B-jk}
\end{align}
By expressing $\ii^2 = \gamma_1$ and $\jj^2 = \gamma_2$, we obtain a commutative hypercomplex algebra $B[\gamma_1,\gamma_2]$ whose multiplication table is
\begin{equation}
\begin{tabular}{c|rrr} 
    $B[\gamma_1,\gamma_2]$ & $\ii$ & $\jj$ & $\kk$ \\ \hline
    $\ii$ & $\gamma_1$ & $\kk$ & $\gamma_1 \jj$ \\  
    $\jj$ & $\kk$ & $\gamma_2$ & $\gamma_2 \ii$ \\  
    $\kk$ & $\gamma_1 \jj$ & $\gamma_2 \ii$ & $\gamma_1\gamma_2$  
\end{tabular} 
\end{equation}
Analogously, we end up with the four-dimensional algebras $B[-1,-1],B[-1,+1],B[+1,-1],B[+1,+1]$ by taking $\gamma_1,\gamma_2 \in \{ -1, +1\}$. Table \ref{tab:non_cd_algebras} contains the multiplication table of these four algebras.

\begin{remark}
Because they have the same multiplication table, the hypercomplex algebra $B[-1,+1]$ corresponds to the tessarines, also known as commutative quaternions \cite{Ortolani2017OnProcessing,Navarro-Moreno2020TessarineCondition}. Similarly, the algebra $B[-1,-1]$ corresponds to the bi-complex numbers \cite{Takahashi2021ComparisonControl} while $B[+1,+1]$ is equivalent to the Klein 4-group, a commutative algebra of great interest in symmetric group theory \cite{Kobayashi2020HopfieldFour-group}. Finally, we would like to point out that the algebras $B[-1,-1]$ and $B[+1,-1]$ can be both obtained from $B[-1,+1]$ by a change of bases. Therefore, the three algebras $B[-1,-1]$, $B[+1,-1]$ and $B[-1,+1]$ are all isomorphic.
\end{remark}

Concluding, we have a total of eight four-dimensional hypercomplex algebras. All of them are associative, four are anticommutative and the remaining are commutative. Furthermore, the hypercomplex units are well structured in their multiplication table. Precisely, their multiplication table can be written as follows
\begin{equation} \label{eq:s_table}
\begin{tabular}{c|rrr} 
          & $\ii$ & $\jj$ & $\kk$ \\ \hline
    $\ii$ & $s_{11}$ & $s_{12}\kk$ & $s_{13} \jj$ \\  
    $\jj$ & $s_{21}\kk$ & $s_{22}$ & $s_{23} \ii$ \\  
    $\kk$ & $s_{31} \jj$ & $s_{32} \ii$ & $s_{33}$  
\end{tabular}
\end{equation}
where $s_{ij} \in \{-1,+1\}$, for all $i,j = 1,\ldots,3$, depends on the parameters $\gamma_1$ and $\gamma_2$ as well as on the commutativity or anticommutativity of the multiplication. The multiplication table \eqref{eq:s_table} helped us to efficiently implement convolutional neural networks based on the eight hypercomplex algebras presented in this section. We detail this remark in the following section.

\section{Hypercomplex-valued Convolutional Neural Networks (HvCNN)} \label{sec:hvcnn}

Convolutional layers are crucial building blocks of convolutional neural networks. The main strength of convolutional layers is their ability to process data locally and, thus, learn local patterns. Convolutional neural networks have been widely applied to image processing tasks \cite{Geron19HandsOn}. Let us briefly describe a real-valued convolutional layer.

\subsection{Real-valued Convolutional Layers}

Suppose a convolutional layer is fed by a real-valued image $\mathbf{I}$ with $C$ feature channels. Let $\mathbf{I}(p,c) \in \R$ denote the intensity of the $c$th channel of the image $\mathbf{I}$ at pixel $p$. The neurons of a convolutional layer are parameterized structures called filters. The filters have the same number $C$ of channels as the image $\mathbf{I}$. Let $D$, commonly a rectangular grid, denote the domain of the filters. Also, let the weights of a convolutional layer with $K$ real-valued filters be arranged in an array $\mathbf{F}$ such that $\mathbf{F}(q,c,k)$ denotes the value of the $c$th channel of the $k$th filter at the point $q \in D$, for $c=1,\ldots,C$ and $k = 1,\ldots,K$. A convolutional layer with $K$ filters yields a real-valued image $\mathbf{J}$ with $K$ feature channels obtained by evaluating an activation function on the addition of a bias term and the convolution of the image $\mathbf{I}$ by each of the $K$ filters. Precisely, let $(\mathbf{I}\ast \mathbf{F})(p,k)$ denote the convolution of the image $\mathbf{I}$ by the $k$th filter at pixel $p$. Intuitively, $(\mathbf{I}\ast \mathbf{F})(p,k)$ is the sum of the multiplication of the weights of the $k$th filter and the intensities of the pixels of the image in a window characterized by the translation of $p$ by $S(q)$, for $q \in D$. The term $S(q)$, for $q$ in the domain $D$ of the filter, represents a translation that can take vertical and horizontal strides into account. In mathematical terms, the convolution of the image $\mathbf{I}$ by the $k$th filter at pixel $p$ is given by
\begin{equation} \label{eq:real-conv}
    (\mathbf{I} \ast \mathbf{F})(p,k) = \sum_{c=1}^C \sum_{q \in D} \mathbf{I}\big(p+S(q),c\big)\mathbf{F}(q,c,k).
\end{equation}
Moreover, the intensity of the $k$th feature channel of the output of a convolutional layer at pixel $p$ is defined by 
\begin{equation} \label{eq:real-conv-layer}
    \mathbf{J}(p,k) = \varphi \left( b(k) + (\mathbf{I}\ast \mathbf{F})(p,k)\right), 
\end{equation}
where $\varphi:\R \to \R$ denotes the activation function.

%
%

\subsection{Hypercomplex-valued Convolutional Layers}

The hypercomplex-valued convolutional layer is defined analogously to the real-valued convolutional layer by replacing the real numbers and operations with their corresponding hypercomplex versions in \eqref{eq:real-conv} and \eqref{eq:real-conv-layer} \cite{Trabelsi17complex,Gaudet2017DeepNetworks}. Precisely, 
the ``intensity'' of the $k$th channel of the hypercomplex-valued output image $\mathbf{J}^{(h)}$ at pixel $p$ is given by 
\begin{equation} \label{eq:hyper-conv-layer}
    \mathbf{J}^{(h)}(p,k) = \varphi_{\A} \left( b^{(h)}(k) + (\mathbf{I}^{(h)} \ast \mathbf{F}^{(h)})(p,k) \right), 
\end{equation}
where $\varphi_{\A}:\A \to \A$ is a hypercomplex-valued activation function, $b_k^{(h)} \in \A$ is the bias term, and
\begin{equation} \label{eq:hyper-conv}
    (\mathbf{I}^{(h)} \ast \mathbf{F}^{(h)})(p,k) = \sum_{c=1}^C \sum_{q \in D}  \mathbf{I}^{(h)}(p+S(q),c)\mathbf{F}^{(h)}(q,c,k), 
\end{equation}
is the convolution of $\mathbf{I}^{(h)}$ by the $k$th hypercomplex-valued filter at pixel $p$. In this paper, we only consider split-functions defined using a real-valued function $\varphi:\R \to \R$ as follows for all $x \in \hyperN{x} \in \A$:
\begin{equation} \label{eq:split-function}
 \varphi_{\A}(x) = \varphi(x_0) + \varphi(x_1)\ii_1+ \ldots + \varphi(x_n)\ii_n.
\end{equation}

\subsection{Emulating Hypercomplex-valued Convolutional Layers} 

Since most deep neural network libraries are designed for real-valued inputs, we show how to emulate a four-dimensional hypercomplex-valued convolutional layer using a real-valued convolutional layer. The reasoning is quite similar to the approaches reported in the literature for complex- and quaternion-valued deep networks \cite{Trabelsi17complex,Gaudet2017DeepNetworks}.

First, an image $\mathbf{I}^{(h)}$ with $C$ channels defined on a four-dimensional hypercomplex algebra can be represented by 
\begin{equation}
    \mathbf{I}^{(h)} = \quat{\mathbf{I}},
\end{equation}
where $\mathbf{I}_0$, $\mathbf{I}_1$, $\mathbf{I}_2$, and $\mathbf{I}_3$ are real-valued images with $C$ channels. Similarly, a bank of $K$ hypercomplex-valued filters can be represented by 
\begin{equation}
    \mathbf{F}^{(h)} = \mathbf{F}_{0} + \mathbf{F}_{1} \ii + \mathbf{F}_{2} \jj + \mathbf{F}_{3} \kk,
\end{equation}
where $\mathbf{F}_{0}$, $\mathbf{F}_{1}$, $\mathbf{F}_{2}$, and $\mathbf{F}_{3}$ are real-valued arrays representing each a bank of $K$ filters with domain $D$ and $C$ feature channels. From the multiplication table \eqref{eq:s_table} and omitting the arguments $(p+S(q),c)$ and $(q,c,k)$ to simplify the notation, we obtain
\begin{align*}
& \mathbf{I}^{(h)}(p+S(q),c)\mathbf{F}^{(h)}(q,c,k) \\
&= (\quat{\mathbf{I}})(\mathbf{F}_{0} + \mathbf{F}_{1} \ii + \mathbf{F}_{2} \jj + \mathbf{F}_{3} \kk) \\
&= \mathbf{I}_0 \mathbf{F}_{0}+ s_{11} \mathbf{I}_1 \mathbf{F}_{1} + 
s_{22} \mathbf{I}_2 \mathbf{F}_{2} + s_{33} \mathbf{I}_3 \mathbf{F}_{3} \\
&\;+(\mathbf{I}_0 \mathbf{F}_{1}+ \mathbf{I}_1 \mathbf{F}_{0} + 
s_{23} \mathbf{I}_2 \mathbf{F}_{3} + s_{32} \mathbf{I}_3 \mathbf{F}_{2})\ii \\
&\;+(\mathbf{I}_0 \mathbf{F}_{2}+ s_{13} \mathbf{I}_1 \mathbf{F}_{3} + 
\mathbf{I}_2 \mathbf{F}_{0} + s_{31} \mathbf{I}_3 \mathbf{F}_{1})\jj \\
&\;+(\mathbf{I}_0 \mathbf{F}_{3}+ s_{12} \mathbf{I}_1 \mathbf{F}_{2} + 
s_{21} \mathbf{I}_2 \mathbf{F}_{1} + \mathbf{I}_3 \mathbf{F}_{0})\kk  
\end{align*}
Therefore, the real-part of the convolution given by \eqref{eq:hyper-conv} satisfies the equation
\begin{align}
     \big(\mathbf{I}^{(h)} \ast \mathbf{F}^{(h)}\big)_0 (p,k) &= \sum_{c=1}^C \sum_{q \in D} \Big[ 
    \mathbf{I}_0(p+S(q),c) \mathbf{F}_{0}(q,c,k) \nonumber \\ 
    & \;+s_{11} \mathbf{I}_1(p+S(q),c) \mathbf{F}_{1}(q,c,k) \label{eq:I*F_0} \\
    & \;+s_{22} \mathbf{I}_2(p+S(q),c) \mathbf{F}_{2}(q,c,k) \nonumber \\
    & \; +s_{33} \mathbf{I}_3(p+S(q),c) \mathbf{F}_{3}(q,c,k)\Big],\nonumber 
\end{align}
Equivalently, the real-part of the convolution of the image $\mathbf{I}^{(h)}$ by the $k$th hypercomplex-valued filter at pixel $p$ can be computed using the real-valued convolution
\begin{equation}
\big(\mathbf{I}^{(h)} \ast \mathbf{F}^{(h)}\big)_0 (p,k) = \big(\mathbf{I}^{(r)} \ast \mathbf{F}_0^{(r)}\big)(p,k),
\end{equation}
where $\mathbf{I}^{(r)}$ is the real-valued image with $4C$ features channels obtained by concatenating the real and imaginary parts of $\mathbf{I}^{(h)}$ as follows 
\begin{equation} \label{eq:I^r}
    \mathbf{I}^{(r)}(p,:) = [\mathbf{I}_0(p,:),\mathbf{I}_1(p,:),\mathbf{I}_2(p,:),\mathbf{I}_3(p,:)],
\end{equation}
for all pixels $p$, and $\mathbf{F}_{0}^{(r)}$ is the real-valued filter defined by
\begin{align}
    & \mathbf{F}^{(r)}_{0}(q,1:C,k) = \mathbf{F}_{0}(q,1:C,k), \\
    & \mathbf{F}^{(r)}_{0}(q,C+1:2C,k) = s_{11}\mathbf{F}_{1}(q,1:C,k), \\ & \mathbf{F}^{(r)}_{0}(q,2C+1:3C,k) = s_{22}\mathbf{F}_{2}(q,1:C,k), \\ & \mathbf{F}^{(r)}_{0}(q,3C+1:4C,k) = s_{33} \mathbf{F}_{3}(q,1:C,k),
\end{align}
for all $q \in D$ and $k=1,\ldots,K$. For short, the notation
\begin{equation}  \label{eq:F_0^r}
    \mathbf{F}^{(r)}_{0} = [\mathbf{F}_{0}, s_{11} \mathbf{F}_{1}, s_{22} \mathbf{F}_{2}, s_{33} \mathbf{F}_{3}],
\end{equation}
means that $\mathbf{F}^{(r)}_{0}$ is obtained by concatenating $\mathbf{F}_0$, $s_{11} \mathbf{F}_{1}$, $s_{22} \mathbf{F}_{2}$, and $s_{33} \mathbf{F}_{3}$ as prescribed above. Note from \eqref{eq:real-conv} that
\begin{align*}
& \big(\mathbf{I}^{(r)} \ast \mathbf{F}_0^{(r)}\big)(p,k) = \sum_{c=1}^{4C} \sum_{q \in D} \mathbf{I}^{(r)}\big(p+S(q),c\big)\mathbf{F}_0^{(r)}(q,c,k) \\ 
& \quad = \sum_{q \in D}\Bigg[ \sum_{c=1}^{C}  \mathbf{I}^{(r)}\big(p+S(q),c\big)\mathbf{F}_0^{(r)}(q,c,k) + \ldots \\
& \qquad + \sum_{c=3C+1}^{4C} \mathbf{I}^{(r)}\big(p+S(q),c\big)\mathbf{F}_0^{(r)}(q,c,k) \Bigg] \\
& \quad = \sum_{q \in D}\Bigg[ \sum_{c=1}^{C}  \mathbf{I}_0\big(p+S(q),c\big)\mathbf{F}_0(q,c,k) + \ldots \\
& \qquad + \sum_{c'=1}^{C} s_{33} \mathbf{I}_3\big(p+S(q),c'\big)\mathbf{F}_3(q,c',k) \Bigg],
\end{align*}
which coincides with $\big(\mathbf{I}^{(h)} \ast \mathbf{F}^{(h)}\big)_0 (p,k)$ given by \eqref{eq:I*F_0}. Therefore, using a split-function, the real-part $\mathbf{J}_0(p,k)$ of the output $\mathbf{J}^{(h)}(p,k)$ of the hypercomplex-valued convolutional layer given by \eqref{eq:hyper-conv} can be computed using a real-valued convolutional layer as follows
\begin{equation}
\mathbf{J}_0(p,k) = \varphi\Big(b_0(k) + \big(\mathbf{I}^{(r)} \ast \mathbf{F}_0^{(r)}\big)(p,k)\big),
\end{equation}
where the bias term $b_0(k)$ corresponds to the real-part of $b(k)$ while $\mathbf{I}^{(r)}$ and $\mathbf{F}_0^{(r)}$ are given by \eqref{eq:I^r} and \eqref{eq:F_0^r}, respectively. In a similar vein, the three imaginary parts $\mathbf{J}_1(p,k)$, $\mathbf{J}_2(p,k)$, and $\mathbf{J}_3(p,k)$ of the $k$th channel of the hypercomplex-valued image $\mathbf{J}^{(h)}$ at pixel $p$ can be computed using real-valued convolutional layers with bias terms $b_{1}(k)$, $b_{2}(k)$, and $b_{3}(k)$ and real-valued filters 
\begin{align}
    & \mathbf{F}^{(r)}_{1} = 
    [\mathbf{F}_{1}, \mathbf{F}_{0}, s_{23} \mathbf{F}_{3}, s_{32} \mathbf{F}_{2}], \\
    & \mathbf{F}^{(r)}_{2} = 
    [\mathbf{F}_{2}, s_{13} \mathbf{F}_{3}, \mathbf{F}_{0}, 
    s_{31} \mathbf{F}_{1}], \\ 
    & \mathbf{F}^{(r)}_{3} = [\mathbf{F}_{3}, s_{12} \mathbf{F}_{2}, 
s_{21} \mathbf{F}_{1}, \mathbf{F}_{0}],
\end{align}
respectively.

We would like to finish this section with few remarks: First, emulating a hypercomplex-valued convolutional layer allow us to take advantage of open-source deep libraries such as \texttt{Tensorflow} and \texttt{PyTorch} for \texttt{python} and \texttt{Flux} for \texttt{Julia Language}. Consequently, hypercomplex-valued versions of well-known deep neural networks can be implemented in current deep learning libraries. On top of that, although not properly designed for hypercomplex-valued models, pooling layers, sophisticated optimizers, and speed-up training techniques such as batch normalization and weight initialization can be incorporated as the first attempt into hypercomplex-valued deep networks. In the following section, we consider a hypercomplex-valued deep neural network for the classification of lymphocytes from smear blood images that resembles the LeNet architecture \cite{LeCun1998Gradient-basedRecognition}.

\section{Lymphoblast Image Classification Task}
\label{sec:Applications}

\begin{figure}
    \centering
    \begin{tabular}{cc}
    \footnotesize{a) Probable lymphoblast} & \footnotesize{b) Healthy cell} \\
    \includegraphics[width=0.4\columnwidth]{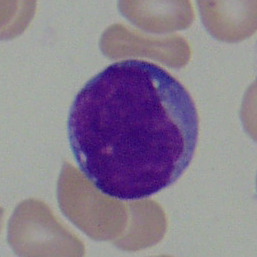} &
    \includegraphics[width=0.4\columnwidth]{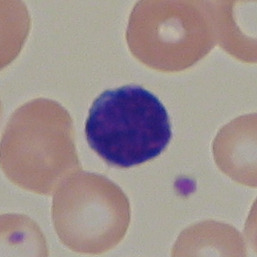}
    \end{tabular}
    \caption{Examples of candidate cells for the classification task. Images from the ALL-IDB dataset \cite{Labati2011All-IDB:Processing}.}
    \label{fig:lymphobast}
\end{figure}

One of the primary forms of diagnosis of acute lymphoblastic leukemia (ALL) is through microscopic blood smear image inspection \cite{Bibi2020IoMT-Based,Genovese2021HistopathologicalDetection,Genovese2021AcuteLearning,Zolfaghari2022ACells}. Precisely, physicians diagnose ALL by the presence of many lymphoblasts in a blood smear image in which white cells are stained with bluish-purple coloration. A candidate cell image is selected, cut from the blood smear image, and fed into a machine learning classifier for a computer-aided leukemia diagnosis. For illustrative purposes, Fig. \ref{fig:lymphobast} shows two candidate cell images used in such a classification task. 
The classifier predicts if the candidate image is a lymphoblast in the simplest binary case. 

In our model, the candidate images have $100 \times 100$ pixels and have been saved using either the RGB (red-green-blue) or the HSV (hue-saturation-value) color encodings. Using the RGB encoding, an image is characterized by the intensities of red (R), green (G), and blue (B), respectively. An RGB encoded image has been mapped into a hypercomplex-valued image $\mathbf{I}^{(h)}_{RGB}$ by means of the equation
\begin{equation}
    \mathbf{I}^{(h)}_{RGB} = R\ii + G\jj + B\kk.
\end{equation}
Similarly, a HSV-encoded image is characterized by the hue $H \in [0,2\pi)$, the value $V \in [0,1]$, and the saturation $S \in [0,1]$. A hypercomplex-valued image $\mathbf{I}^{(h)}_{HSV}$ is derived from an HSV-encoded image as follows \cite{Granero2021Quaternion-ValuedDiagnosis}:
\begin{equation} \label{eq:hyper-HSV}
    \mathbf{I}^{(h)}_{HSV} = \big(\cos(H) + \sin(H) \ii\big)(S + V\jj).
\end{equation}
Note that, because $\ii \jj = \kk$ in all the considered hypercomplex algebras, \eqref{eq:hyper-HSV} is equivalent to 
\begin{equation}
 \mathbf{I}^{(h)}_{HSV} = S \cos(H) + S \sin(H)\ii + V \cos(H) \jj + V \sin(H)\kk.
\end{equation}

We performed the lymphocyte classification task using real-valued CNNs (RvCNNs) and the proposed HvCNNs. Similar architectures are adopted for both real- and hypercomplex-valued models, as suggested in \cite{Granero2021Quaternion-ValuedDiagnosis}. The RvCNN features three convolutional layers with $3\times 3$ filters followed by a max-pooling layer with $2\times2$ kernels. A dense layer with 1 unit yields the output. The hypercomplex-valued models have the same layer layout but a much smaller number of filters per convolution layer because each hypercomplex-valued channel is equivalent to four real-valued feature channels. The activation function adopted for all convolutional layers is the rectified linear unit (ReLU). The dense layer for both architectures is a single neuron that outputs the label $0$ for a healthy white cell and $1$ for a lymphoblast image. Table \ref{tab:params} shows a breakdown of total parameters per layer for each architecture. All deep network models in this work were implemented using the python libraries \texttt{Keras} and \texttt{Tensorflow}. The source codes are available at \url{https://github.com/mevalle/Hypercomplex-valued-Convolutional-Neural-Networks}.

\begin{table}
\centering
    \caption{Parameter distribution per layer for each architecture.}
    \label{tab:params}
\begin{tabular}{|c|c|c|c|} \hline
    & & RvCNN & HvCNN \\ \hline
    \multirow{2}*{Conv Layer 1} & (3,3) filters & {32} & {8} \\ \cline{2-4}
    & Parameters & {896} & {320} \\ \hline
    \multirow{2}*{Conv Layer 2} & (3,3) filters & {64} & {16} \\ \cline{2-4}
    & Parameters & {18,496} & {4,672} \\ \hline
    \multirow{2}*{Conv Layer 3} & (3,3) filters & {128} & {32} \\ \cline{2-4}
    & Parameters & {73,856} & {18,560} \\ \hline
    \multirow{2}*{Dense Layer} & Units & 1 & 1 \\ \cline{2-4}
    & Parameters & 12,801 & 12,801 \\ \hline
    \textbf{Total} & & 106,049 & 36,353 \\ \hline
    \end{tabular}    
\end{table}

\subsection{Computational Experiments}

Let us now describe the outcome of the computational experiments performed using the acute lymphoblastic leukemia image database (ALL-IDB), a popular public dataset for segmentation and classification tasks directed at ALL detection \cite{Labati2011All-IDB:Processing}. The ALL-image database (ALL-IDB) consists of two sets of images. The ALL-IDB1 contains 108 blood smear images for segmentation and classification tasks. The ALL-IDB2 contains 260 segmented images, each containing a single blood element like the images depicted in Fig. \ref{fig:lymphobast}, and is aimed exclusively for the classification task \cite{Labati2011All-IDB:Processing}.

We used all the 260 images from the ALL-IDB2 dataset in our computational experiments. Like Genovese et al. \cite{Genovese2021HistopathologicalDetection,Genovese2021AcuteLearning}, we randomly split the dataset into the training and test sets containing each 50\% of the total number of images. The training set is enlarged using horizontal and vertical flips. We conducted 100 simulations per experiment. Each simulation consists of splitting training and test set, augmenting the training set, initializing the network parameters, training for 100 epochs using ADAM optimizer and the binary cross-entropy loss function, and predicting test images. We evaluate the performances using the accuracy score in the test set.

\begin{figure*}
    \centering
    \begin{tabular}{cc}
    \footnotesize{a) RGB-encoded images}  & \footnotesize{b) HSV-encoded images} \\
    \includegraphics[width=0.48\textwidth]{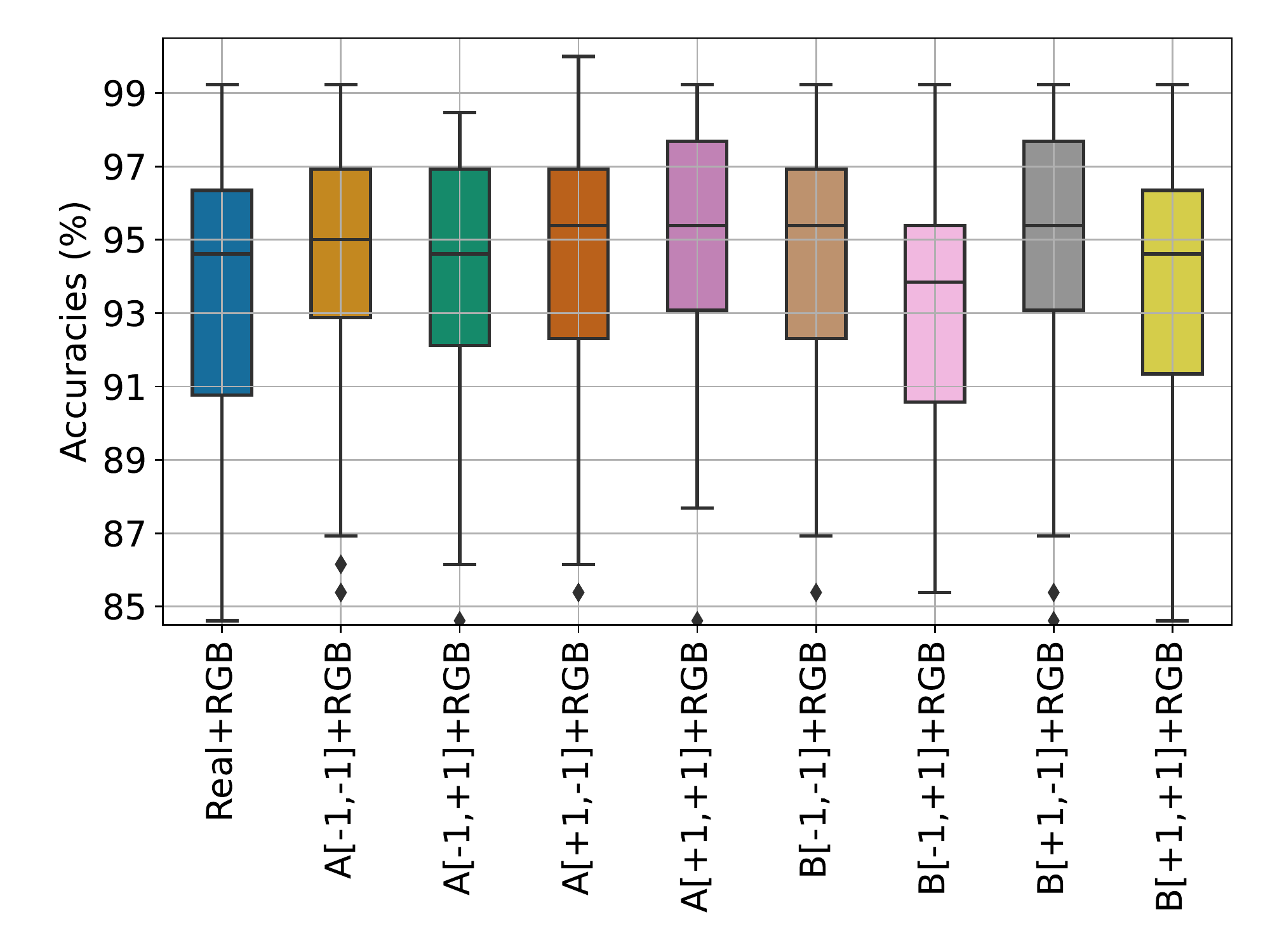}  &
    \includegraphics[width=0.48\textwidth]{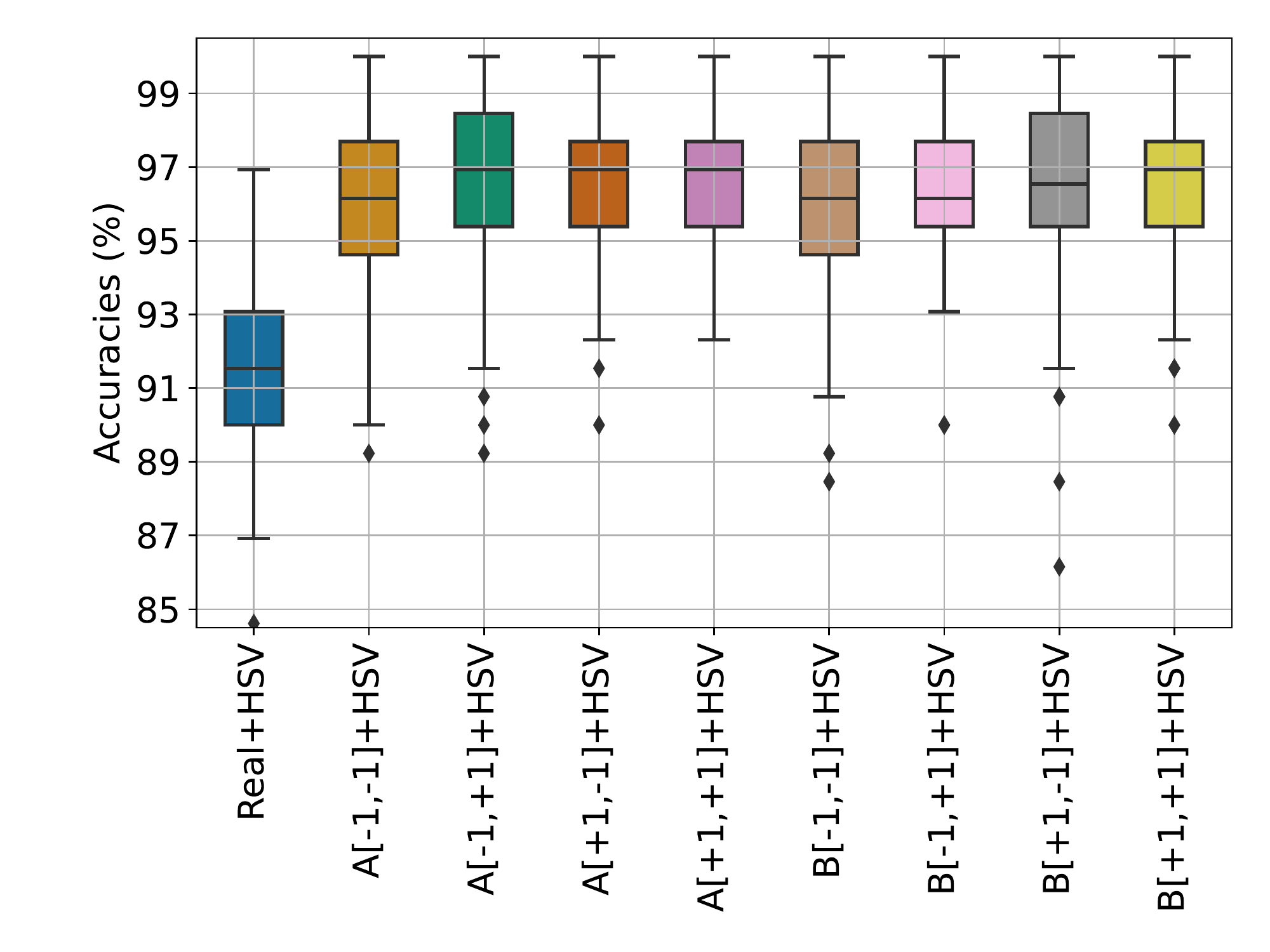}
    \end{tabular}
    \caption{Boxplot of test set accuracies produced by real-valued and hypercomplex-valued neural networks.}
    \label{fig:boxplot}
\end{figure*}

We derive a total of 18 network configurations using the real numbers and the eight hypercomplex algebras detailed in Section \ref{sec:hc-matrix-algebra}. Namely, each of the nine models (1 RvCNN and 8 HvCNNs) is used in the classification tasks of both RGB and HSV encoded image sets. The box plots depicted in Fig. \ref{fig:boxplot} summarize the outcome of the computational experiment. 

Note that all models performed well for the RGB encoded images, with medians close to $95\%$ accuracy except for the HvCNN based on the algebra $B[-1,+1]$ that yields a median accuracy rate of $93.8\%$. In the HSV case, however, performances vary more drastically. The real-valued model exhibits poor performance compared to all the hypercomplex-valued ones, with a median accuracy rate of $91.5\%$. The HvCNNs based on the algebras $A[-1,-1]$, $B[-1,-1]$, and $B[-1,+1]$ yielded all a median accuracy score of $96.2\%$ while the HvCNN based on the algebra $B[+1,-1]$ achieved a median accuracy score of $96.5\%$. Moreover, the HvCNNs based on the isomorphic algebras $A[-1,+1]$, $A[+1,-1]$, and $A[+1,+1]$ as well as the HvCNN based on $B[+1,+1]$ yielded all a median accuracy score of $96.9\%$. The significant improvement in the accuracy scores of the HvCNN models indicates they can take advantage of the locally cohesive structure of the HSV encoding.

\begin{figure}
    \hspace{-1.5em}
    \includegraphics[width=1.1\columnwidth]{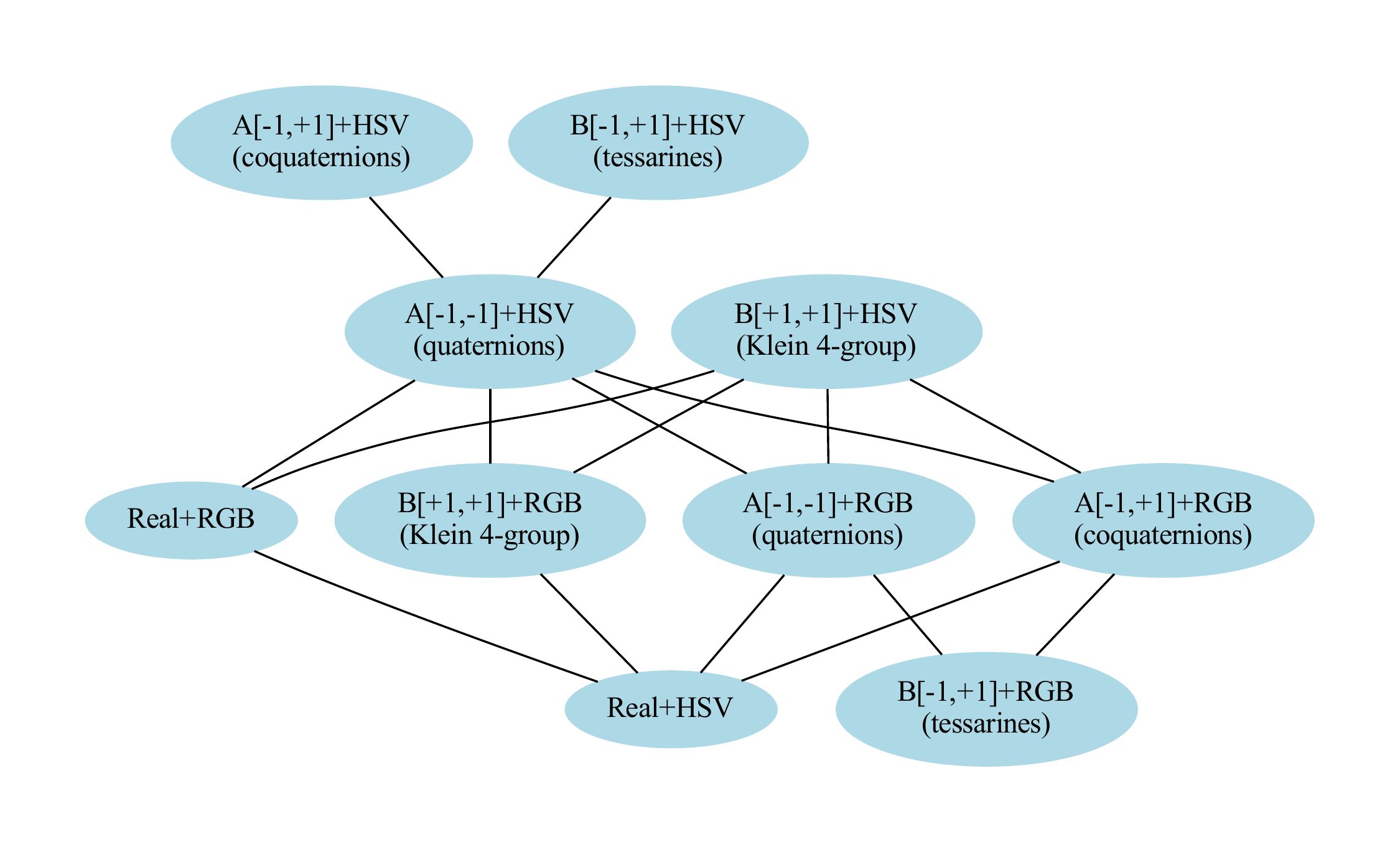}
    \caption{Hasse diagram of representative experiment configurations.}
    \label{fig:hot}
\end{figure}
To better depict the outcome of this computational experiment, we summarize the performance of the network classifiers in the Hasse diagram shown in Fig. \ref{fig:hot}. This figure depicts a single HvCNN model of each isomorphic hypercomplex algebra group to simplify the exposition. Precisely, Fig. \ref{fig:hot} compares the performance of the HvCNNs based on the hypercomplex algebras $A[-1,-1]$ (quaternions), $A[-1,+1]$ (coquaternions), $B[-1,+1]$ (tessarines), $B[+1,+1]$ (Klein 4-group), along with the real-valued models. In this diagram, a solid line connecting two models indicates that the one above achieved better performance than the one below, with a confidence level of 0.99 according to a Student's t-test. In other words, models higher up in the diagram perform significantly better than those on the lower end. Furthermore, the solid lines indicate a transitive relation. Thus, if model A is better than B and B is better than C, then A is better than C. First off, Fig.  \ref{fig:hot} confirms that the HvCNN models performed better using the HSV than the RGB encoding, and HvCNNs on HSV encoded images outperform the real-valued models on both encodings. In addition, it shows the real-valued model on HSV encoded images as the poorest performer. Moreover, coquaternion- and tessarine-based HvCNNs outperformed the HvCNN based on quaternions, the four-dimensional algebra most widely used in applications. At this point, we would like to recall that superior performance of neural networks based on non-usual hypercomplex algebras has been previously reported in the literature \cite{Vieira2020ExtremeAuto-Encoding,Vieira2022AMachines}, despite quaternion-based neural network yielding better performance in applications like controlling a robot manipulator \cite{Takahashi2021ComparisonControl}. 

Finally, we would like to recall that Genovese et al. obtained average accuracy rates of $97.92\%$ using a ResNet18 combined with histopathological transfer learning \cite{Genovese2021HistopathologicalDetection}. The top-performing  HvCNNs achieved average accuracy scores of $96.51\%$ and $96.39\%$. However, in contrast to the ResNet18 network with approximately 11.4M parameters, the HvCNNs have only 36K trainable parameters. In particular, the coquaternion-valued HvCNN with HSV encoding achieved $98\%$ of the average accuracy score of the ResNet18 with transfer learning but with only $0.3\%$ of its total number of trainable parameters.

\section{Concluding Remarks} \label{sec:concluding-remarks}

In this work, we extended the concept of a quaternion-valued CNN to general hypercomplex algebras. We constructed eight such algebras with desirable properties according to Kantor and Solodovnikov framework. These algebras are isomorphic to well-known four-dimensional algebras: quaternions, coquaternions, tessarines, Klein 4-group, Cayley-Dickson algebras, and Clifford algebras. The hypercomplex-valued neural networks have been applied for lymphoblast image classification.

Using the public dataset ALL-IDB2, we conducted experiments featuring real-valued and hypercomplex-valued models. We observed that the HvCNNs performed similarly to the real-valued models in the RGB-encoded images while outperforming the RvCNNs with HSV-encoded images. The superior performance of the HvCNN indicates that they take better advantage of the HSV color system, which is more akin to human vision and more widely used in computer vision applications \cite{cheng2001color}. Consequently, HvCNN models seem more efficient at comprising information in its filters and four-dimensional entities than the real-valued models. Moreover, coquaternion- and tessarine-valued models outperformed the quaternion-valued CNNs. Lastly, the performance attained by the top-performing HvCNN models is significant and comparable to notable results in the literature. Indeed, Genovese et al. observed an average accuracy score of $97.92\%$ with the ResNet18, a deep network containing around 11 million parameters \cite{Genovese2021HistopathologicalDetection}. The coquaternion-valued CNN yielded an average accuracy of $96.51\%$ with approximately $0.3\%$ of the total of trainable parameters of the real-valued ResNet18.



\end{document}